\documentclass{article}
\usepackage{ijcai22}
\usepackage{tabularx}
\usepackage{times}
\usepackage{soul}
\usepackage{url}
\usepackage{multirow}
\usepackage[hidelinks]{hyperref}
\usepackage[utf8]{inputenc}
\usepackage[small]{caption}
\usepackage{subcaption}
\usepackage{graphicx}
\graphicspath{ {./Figures/} }
\usepackage[dvipsnames]{xcolor}
\usepackage{amsmath}
\usepackage{amssymb}
\usepackage{amsthm}
\usepackage{float}
\usepackage{booktabs}
\usepackage{algorithm}
\usepackage{algorithmic}
\usepackage[normalem]{ulem}
\newcounter{notecounter}
\newcommand{\enotesoff}{\long\gdef\enote##1##2{}}

\enotesoff

\newcommand\yzs{\bgroup\markoverwith{\textcolor{red}{\rule[0.5ex]{2pt}{0.4pt}}}\ULon}

\newcommand{\yz}[1]{\textcolor{black}{#1}}

\theoremstyle{definition}
\newtheorem{definition}{Definition}[section]

\title{Behaviour-Diverse Automatic Penetration Testing: A Curiosity-Driven Multi-Objective Deep Reinforcement Learning Approach}

\author{
Yizhou Yang$^1$
\and
Xin Liu$^{1,2}$\and
\affiliations
$^1$Zhejiang Lab\\
$^2$National Research Centre of Parallel Computer Engineering and Technology\\
\emails
yizhou.yang@zhejianglab.com
}

\begin{document}

\maketitle

\begin{abstract}

Penetration Testing plays a critical role in evaluating the security of a target network by emulating real active adversaries. Deep Reinforcement Learning (RL) is seen as a promising solution to automating the process of penetration tests by reducing human effort and improving reliability. Existing RL solutions focus on finding a specific attack path to impact the target hosts. However, in reality, a diverse range of attack variations are needed to provide comprehensive assessments of the target network's security level. Hence, the attack agents must consider multiple objectives when penetrating the network. Nevertheless, this challenge is not adequately addressed in the existing literature.
To this end, we formulate the automatic penetration testing in the Multi-Objective Reinforcement Learning (MORL) framework and propose a Chebyshev decomposition critic to find diverse adversary strategies that balance different objectives in the penetration test. 
Additionally, the number of available actions increases with the agent consistently probing the target network, making the training process intractable in many practical situations. Thus, we introduce a coverage-based masking mechanism that reduces attention on previously selected actions to help the agent adapt to future exploration. Experimental evaluation on a range of scenarios demonstrates the superiority of our proposed approach when compared to adapted algorithms in terms of multi-objective learning and performance efficiency.

\end{abstract}

\section{Introduction}

Penetration testing, shortly pen-testing, is an effective method that aims to assess the vulnerabilities in a cyber system by simulating the activities of an actual attacker. Whereas current pen-testing requires costly and extensive manual effort, especially for large and complex networks, the outcomes are primarily dependent on the experience and knowledge of the pen-testers, which vastly reduce the repeatability. Therefore, to increase consistency, efficiency and reduce cost, time, and human intervention in network assessment, the need for automatic pen-testing is paramount. Such automated pen-testing agents should be able to intelligently reason about future plans based on the uncertain state of the target network environment and autonomously generate attack plans that expose the target network system’s vulnerabilities. However, \yz{leveraging automation} in realistic pen-testing scenarios encounters various hurdles, including partial network observability and scalability to exponentially expanding network dimensions. These problems limit the deployment of automatic pen-testing by using conventional planning algorithms prohibitive \cite{hoffmann2015simulated}. 

Recently, RL has demonstrated \yz{human-level} performance on varies applications, such as Chess~\cite{silver2017mastering}, StarCraft~\cite{vinyals2019grandmaster} and robotics~\cite{laskin2020reinforcement}. These successes led to a proliferation of studies that attempt to apply RL in pen-testing \cite{Hu2020}, \cite{zhou2021autonomous}, \cite{tran2021deep}, \cite{schwartz2019autonomous}.
The search space of attack paths is affected by the network scale, thereby limiting the learning efficiency of the agent. Moreover in actual pen-testing, the attack agent starts with limited information of the target network and will be looking for any foothold they can leverage to gain access into a network. In other word, each action has its pre-condition, which is the state conditions that must be satisfied if the action is to be available. 
Therefore, the action space of the agent is dynamically increasing and hence reduces the tractability of the training process. \yz{This particular challenge} of changing action spaces is not examined as in most RL researches the all the available actions are initially given to the agent. 
These approaches have focused on single-objective RL settings where the agent will learn the optimal attack path that has maximized rewards, which result in monotonous behaviors. In practice, to build sophisticated defenders, it is always essential to have diverse behavioural attack strategies to train against. The diversity of adversary strategies of pen-testing comes from the trade-off between maximizing exploits and vulnerability assessments and minimizing the time cost used to impact the target hosts. As these objectives \yz{maybe} complementary and conflicting \yz{in practice}, it is typically unclear how to evaluate available trade-offs between different objectives in advance. Therefore, it is desirable to produce a set of strategies containing non-dominated solutions.

In the paper, we frame pen-testing as a Multi-Objective Markov Decision Process (MOMDP) with imperfect information of the network system and expanding action space involving scan and exploitation of discovered nodes, which represents a more practical setting than considered in most prior research. 
We opt for a model-free Proximal Policy Optimization (PPO) algorithm conjugated with Random Network Distillation (RND) to train the pen-testing agent with sparse environment rewards. To address the challenge of growing action space in RL training, we propose a coverage masking mechanism that imitates real pen-testers who tend to pay more attention to the newly explored part of the target network. At the same time, in order to provide comprehensive assessments on the security of the target network, we propose a Chebyshev critic to produce diverse attack strategies without prior manual preference settings.

To the authors' best knowledge, we are the first to propose: a highly efficient and stable RL-based automated pen-testing agent with diversified strategies, satisfying the special requirements of pen-testing in real and complex scenarios.

\begin{figure*}[t]
  \begin{subfigure}{0.6\textwidth}
    \includegraphics[width=1.11\textwidth]{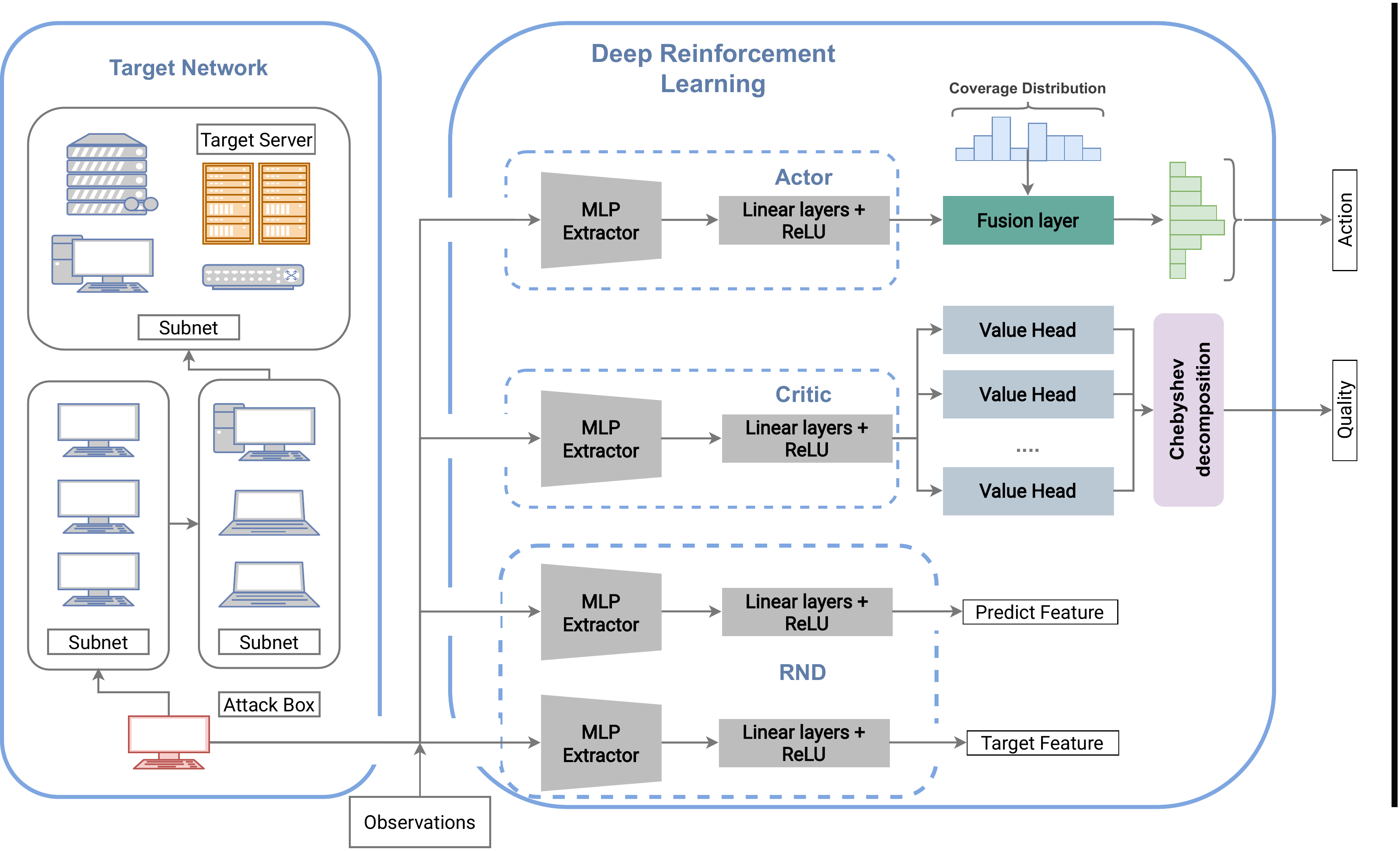}
    \caption{System Architecture} \label{fig:1a}
  \end{subfigure}%
  \hspace*{\fill}   
  \begin{subfigure}{0.31\textwidth}
    \includegraphics[width=0.87\linewidth]{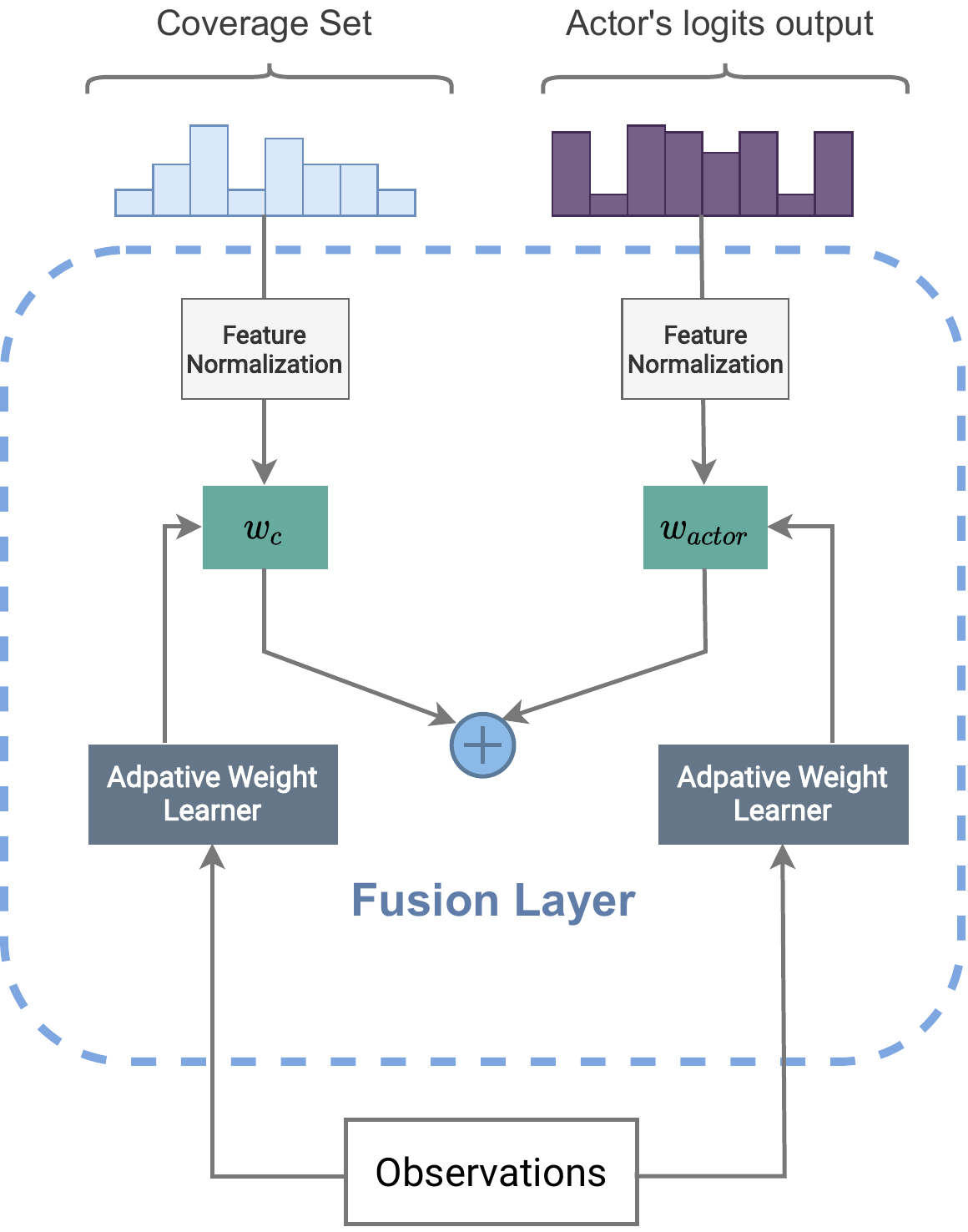}
    \caption{Fusion Layer} \label{fig:1b}
  \end{subfigure}%
\caption{Architecture of our proposed method. Attack agent's observations are fed through separate MLP extractors to Actor-Critic Network and RND respectively. The coverage distribution is fused with logits output of the actor network.} \label{fig:sys_config}
\end{figure*}

\section{Related Works}
Early research focused on modelling PT planning as attack
graphs and decision trees which reflected the view of PT practice as sequential decision making. 
Amongst the most significant contributions, attack graphs were used to represent atomic components (attack actions) by a conjunctive pre-condition and post-condition over relevant properties of the system under attack \cite{applebaum2016intelligent}. This approach was more related to classical planning and attempted to find the optimized attack graph. Despite these approaches offer interpretable and formal models, such algorithms are only valid for small to medium sized networks with fewer than 20 hosts \cite{lallie2020review}, and therefore are extremely hard to scale to large network topolgies. 



Recently, reinforcement learning algorithms, such as tabular Q-learning \cite{massimo2020modeling} and \yz{D}eep Q-network (DQN) \cite{schwartz2019autonomous}, were applied to solve the pen-testing problems where \yz{it} was modeled as a Markov Decision Process (MDP) or a Partially Observable Markov Decision Process (POMDP). \yz{In \cite{tran2021deep}, Hierarchy DQN is proposed to improve the training efficiency, by decomposing the large discrete action space into manageable sets for separate DQN agents. In the same vein, Improved DQN \cite{zhou2021autonomous} decoupled the action into a vector and splits the DQN network into separated streams to estimate the value of each elements in the action vector. Additionally, Intrinsic Curiosity Module (ICM) was also adopted in \cite{zhou2021autonomous} to address the challenge of spare rewards. } 
Meanwhile, \cite{Hu2020} identified the efficient attack path to exploit from the possible solutions generated from traditional search algorithms for vulnerabilities assessments in particular subnet. Together these studies provide important insights into adopting RL for autonomous pen-testing in large scale networks. Extending DRL to networks with a larger action space is still a standing challenge not only in cybersecurity setting but also in RL context \cite{schwartz2019autonomous}.


\cite{vezhnevets2017feudal} presented a hierarchy method to solve large complex space, however finding right abstractions requires sufficient domain knowledge. By mixing two policies defined action space on different levels of granularity, \cite{czarnecki2018mix} proposed a curriculum over agents. \textcolor{black}{In this case, the curriculum prescribes a sampling strategy for the dimensions in order to guide exploration.} \cite{farquhar2020growing} introduced the concept of growing action spaces (GAS), which successively increases the action space from few using a growth schedule. Although the action space is dynamic as well, it differs from our setting in that the action space is expanded as a result of penetration.


Within the literature, pen-testing is usually formalized \yz{as} single-objective \yz{reinforcement learning task} where the goal of agent is to impact the target hosts. A straightforward and effective multi-objective optimization method is Decomposition, which explicitly decompose the multi-objective problem into a set of scalar optimization subproblems. Solving each scalar optimization subproblem usually leads to a Pareto optimal solution. Decomposition has fostered a number of researches, e.g., Cellular-based MOGA \cite{murata2001specification} and NSGA-III \cite{deb2013evolutionary}. Such idea has also been adopted in MORL\cite{hsu2018monas} \cite{mossalam2016multi} as series of single-objective problems with combinations of the scalarization weights. The league-based methods, \cite{jaderberg2019human} train multiple agents simultaneously to explore strategy space and require massive computation resources. However, it is impractical to use league-based methods in pen-testing, as it suffers from heavy dependence of experts' domain knowledge and substantial labor costs \cite{shen2021generating}. 





\section{Preliminary}
\subsection{Multi-Objective MDP}
In single-objective reinforcement learning, the goal of the agent is to find a mapping from states to actions that maximizes the total scalar reward received from the environment. To address the research challenge in this study: \textit{how to choose the best action in a given scenario to ultimately achieve some pre-determined objectives for red agents in penetration testing?} the penetration testing process is modelled as an MOMDP, defined by the tuple $<\mathcal{S}, \mathcal{A}, \mathcal{P}, \overrightarrow{\mathcal{R}}, \gamma>$, corresponding to state space $\mathcal{S}$, action space $\mathcal{A}$, state transition probability $\mathcal{P}$, reward function $\overrightarrow{\mathcal{R}}: \mathcal{S} \times \mathcal{A} \times \mathcal{S} \rightarrow \mathbb{R}^{n}$, 
and discount factor $\gamma \in [0,1]$ used to determine the importance of long term rewards. Noted that instead of scalar reward $\mathcal{R}$ as in SORL, in MOMDP the reward $\overrightarrow{\mathcal{R}}$ is a vector with $n$ individual reward, each corresponding to an objective,.

\paragraph{State} The state is a matrix representation of the attack agent's observation of the network, which should contain the topology of the penetrated part of the target network as well as the information of known hosts. 

\paragraph{Action} The action space represents abstracted operations that an adversary can take within a network, \yz{and each action is corresponding to a high-level adversary behavior in}  Adversarial Tactics, Techniques, and Common Knowledge (ATT\&CK) framework \cite{strom2018mitre}. \textcolor{black}{Each action may be performed on a host (e.g. Exploit), and each host-to-host action (e.g. Service Scan) or host-to-subnet (e.g. Subnet IP Scan) action may target another host or subnet, respectively.} 

\paragraph{Rewards}
In MOMDP, the reward function received by the agent is a vector of $n$ rewards, i.e. one per objective:
\begin{equation}
    \overrightarrow{\mathcal{R}}(s_t, a_t) = \mathbf{r}_t = (r_1(s_t, a_t), r_2(s_t,a_t)...r_n(s_t,a_t)).
\end{equation}
\yz{where $r_i$ is the reward corresponding to objective $i$. Whereas, in conventional SORL, reward shaping \cite{oh2021creating} that adding objective-related reward items in scalar reward $R(s, a)$ is used to learn desirable behaviors. Such that, the reward is shaped as $R(s, a)=\sum_{w_i \in w} w_i *r_i(s, a),$ where weight $w_i$ is associate with objective $i$. Consequently, adjusting weights contributes to guiding the agent towards desirable behaviors.}


\subsection{Proximal Policy Optimization}
Proximal Policy Optimization (PPO), a state-of-the-art model-free on-policy algorithm is used as our learning scheme, which responds more stably to changes in the environment as it can determine the optimal step size to update the gradient. The objective function of PPO is as follows:
\begin{equation}
J^{\textrm{CLIP}}(\theta)=\hat{\mathbb{E}}_{t}\left[\min \left(\textrm{clip}\left(p_{t}(\theta), 1-\epsilon, 1+\epsilon\right) \hat{A}_{t}, p_{t}(\theta) \hat{A}_{t}\right)\right],
\end{equation}\label{eq:clip}
where $p_{t}(\theta)$ is the probability ratio between old policy and new policy. Generalized Advantage Estimator (GAE) \cite{schulman2015high} is used to compute the advantage function $\hat{A}_t$:
\begin{equation}
\hat{A}_t = \sum_{l=0}^{\infty}(\gamma \lambda)^{l} \delta_{t+l}^{V},
\label{eq:GAE}
\end{equation}
where $\gamma$ and $\lambda$ perform a bias-variance trade off of the trajectories. 

As was mentioned above, in penetration testing scenarios, the rewards received by the agent from the target network are very sparse of large value, which can make the RL algorithm completely fail to learn. Therefore, to incentivize exploration \yz{for RL agents in such spares reward settings, Random Network Distillation (RND) is also used to provide curiosity rewards that drive the attack agent to explore the target network.}

\section{Method Framework}

The following section describes how we solve: 
1) the challenge of diversifying attack strategies to balance multi-objectives in pen-testing;
2) the issue of increasingly large discrete action space using innovative deep reinforcement learning framework, namely CLAP. This framework contains three primary components: Deep reinforcement learning (DRL), RND predictor, and RND target. DRL is used to train the output policy of the agent, and for the intrinsic and extrinsic values. The RND predictor and target generate a feature space each, and the exploration bonus is obtained by calculating the difference between the two spaces. 

\yz{Our proposed method adopts the actor-critic framework which employs different multi-layer perceptrons (MLP) for different purposes as shown in Figure~\ref{fig:sys_config}. }All the MLP feature extractors has two layers, each with 256 neurons. For the separate critic value streams each of the fully connected (FC) layers are of size 128. The adaptive weight learners have FC layers of size 256*128. 

The training procedure of our method is described in Algorithm~\ref{alg:algorithm} and will be further explained in the following sections. 



\subsection{Coverage Mask Mechanism}

\textbf{Repeat again: Expanding Action Space}

This mechanism works as human penetration testing experts in realistic scenarios; after successfully pivot to a new subnet or a host, the expert will focus more on gathering information from the new discoveries \yz{and exploiting those information}. \yz{We introduce a} coverage set \yz{that should be maintained during training} to keep track of what actions has been used in the  \yz{past}. 


\yz{The} coverage vector tracks the attention history of  previous iteration\yz{s} and then adjusts the attention of the current iteration, which encourages the actor network to generate action that concentrates more on the recent discovered part of the target network, reducing repetitive operations on penetrated subnets and hosts. 

In the training phase, at step $t$, the coverage vector is the sum of $\mathbb{A}_c^t$ which \yz{is} the one-hot embed\yz{ding}  of the action taken by the agent at time $t$ \footnote{ Actions that haven't been revealed yet are padded with zeros}:

\begin{equation}
    C_t=\sum^{t-1}_{t^{'}=0} \mathbb{A}_c^t.
\end{equation}

To integrate the coverage vector that represents historical action selections and the actor's output that represents the current probability of action selection. The historical attention can therefore be informed to the actor. We proposed a fusion integration method that assigns adaptive fusion weights to the coverage vector and actor's output, as shown in Figure~\ref{fig:1b}.

We enforce the fusion weights $W_c$ and $W_{\operatorname{actor}}$ to explicitly depend on the observation (state), i.e. $W_c = \Phi_c (s), W_{\operatorname{actor}} = \Phi_{\operatorname{actor}}(s)$. Different attack agent's observation of the target network would lead to modifications to the adaptive weight learner dynamically. As a consequence, the the fusion layer can fast adapt to the pen-testing process and favourably learn proper fusion weights. A auxiliary loss is also defined, to promote the learning process of our proposed fusion layer and reduce repetitive operations on the same subnet or host:

\begin{equation}
    L^{\operatorname{Coverage}}=\sum_{i} \min \left(a_{i}^{t}, c_{i}^{t}\right)
\end{equation}

\subsection{Weighted Chebyshev Critic Scalarization}

In MOMDP, since each of the components in the reward vector represents a different objective when optimizing towards one or more goals, conflicts will likely arise. Nevertheless, as implied in Sutton's reward hypothesis \cite{sutton2018reinforcement}, 
MOMDPs can also be converted into single-objective MDP by converting the vector-valued reward function to a scalar reward. Hence the problem may be solvable with single-objective methods, such conversion leads to the following definition \cite{roijers2013survey}:

\begin{definition}
A scalarization function $f$, is a function that projects the multi-objective value $V$ to a scalar value.
\begin{equation}
    V_{\mathbf{w}}(s)=f\left(\mathbf{V}(s), \mathbf{w}\right),
\end{equation}
where $\mathbf{w}=(w_1, w_2...w_n)$ is the vector of preference weights parameterizing $f$.
\end{definition}
This scalarization function maps each possible policy value vector, $V$ onto a scalar value. In this study, we consider \textit{unknown weights scenario} in \cite{roijers2013survey}. To this end, trade-offs between these objectives lead to a set policies for each possible preference that a user might have.

The most commonly used scalatization function is the linearly weighted sum $f\left(\mathbf{V}, \mathbf{w}\right)=\mathbf{w} \cdot \mathbf{V}$, notwithstanding this method has a fundamental limitation as it can only find policies that lie in convex regions of the objective function. The policy optimization process of the each objectives in pen-testing are correlated to the topology of the network, which means the shape of the optimal set may have several local concavities. To address this issue, we therefore propose to use the $L_p$ metric:


\begin{equation}
L_{p}(x)=\left(\sum_{o=1}^{n} w_{o}\left|f_{o}(x)-z_{o}^{*}\right|^{p}\right)^{1 / p}, 
\end{equation}
where $1 \leq p \leq \infty$ and $w_o$ is user’s relative preference for the $o$ objectives. Specifically,  $L_p$ metric measures the distance between a point in the multi-objective space and a utopian point $z^{*}$. In the case of $p=\infty$, the metric is also call the Chebyshev metric:
\begin{equation}
    L_{\infty}(x)=\max _{o=1 \ldots n} w_{o}\left|f_{o}(x)-z_{o}^{*}\right|.
\end{equation}

Therefore, the value estimated by the critic is obtained by using Chebysheb metric:

\begin{equation}
    V_{\operatorname{cheby}}(S) = \max_{o = 1,...n} w_o \cdot |V_o(S;\omega) - Z_o^*|.
    \label{eq:v_update}
\end{equation}

It should be noted that, the referenced value $Z_o^*$ it obtained by keep recording the best value so far for each objective plus a small offset. Under this settings, when the actor-critic update its policy network, the value of critic is replaced by the Chebyshev scalarization function. As a result, at step $t$, $\delta_{t+l}^{V}$ at the core of the policy update Equation~\eqref{eq:GAE} can be expressed as follow:

\begin{equation}
\delta_{t}^{V} = r_t + \gamma V_{\operatorname{cheby}} (S_{t+1}; \omega) - V_{\operatorname{cheby}}(S_{t}; \omega),
\end{equation} \label{eq:v_update2}

\begin{algorithm}[htb]
\caption{Training Process of CLAP}
\label{alg:algorithm}
\textbf{Input}: Initialize all the networks
\begin{algorithmic}[1] 
\FOR{Episode $e=0,1,2...$}
\STATE Initialize coverage vector $c$
\WHILE{$s_t \neq $ terminated}
\STATE select an action $a_t$ following policy $\pi(\cdot \vert s_t, c_t ;\theta, \phi )$
\STATE Execute action $a_t$ and obtain reward $r_t$ and new observation $s_{t+1}$
\STATE Encode action $a_t$ and update $c$
\STATE Store this transaction in the memory buffer $M_{\operatorname{PPO}}$ and $M_{\operatorname{RND}}$
\STATE Using GAE to compute the internal advantage to go $\hat{A}^{int}_t$ from the predictor model $\hat{f}(\beta)$ and feature model $f(\beta)$.
\STATE Using GAE to compute the external advantage to go $\hat{A}_t^{\operatorname{ext}}$ using Equation~\eqref{eq:v_update2}
\STATE Calculate the total advantage to go and update the policy by maximizing the PPO-clip objectives Equation~\eqref{eq:clip}
\STATE Update the target networks

\ENDWHILE
\ENDFOR
\STATE \textbf{return} solution
\end{algorithmic}
\end{algorithm}


Alternatively, this procedure can be applied to estimate the entire non-inferior tradeoff among the objectives by iterating through the full range of values for the weight vector.

\section{Performance Evaluation}

\yz{In the following sections, we will perform an empirical analysis on two of benchmark environments: Network Attack Simulator (NASim) and modified variants of 2021 IJCAI-21 ACD CAGE-Challenge \cite{cage_challenge_1}. }


\subsection{Benchmark Scenarios}
\paragraph{NASim}
NASim is an open-source platform that provides varies abstract network scenarios for testing pen-testing agents using the RL algorithms. As the reward of the agent is the value of compromised hosts minus the cost of the actions. The objective of the agent is to compromise all target hosts with positive value on the network while minimizing the number or cost of actions used.  Available actions in NASim include scan (gathering information of hosts/subnets), exploit (exploit vulnerable services), and privilege escalation (using processes to elevate the access). The size of the action space in NASim $|A|$, is calculated as $|A| = |H| \times (N^{\operatorname{Exploits}} + N^{\operatorname{PrivEscs}} + N^{\operatorname{Scans}})$, where $H$ is the set of hosts. $N^{\operatorname{Exploits}}$, $N^{\operatorname{PrivEscs}}$ and $N^{\operatorname{Scnas}}$ denote the number of available types of exploits, privilege escalations and scans of each host, respectively. 
\begin{figure}[htb!]

\centering

\subfloat[Small Network Scenario (left: Number of actions, right: Rewards)]{
\includegraphics[width=0.5\linewidth, height=0.32\linewidth]{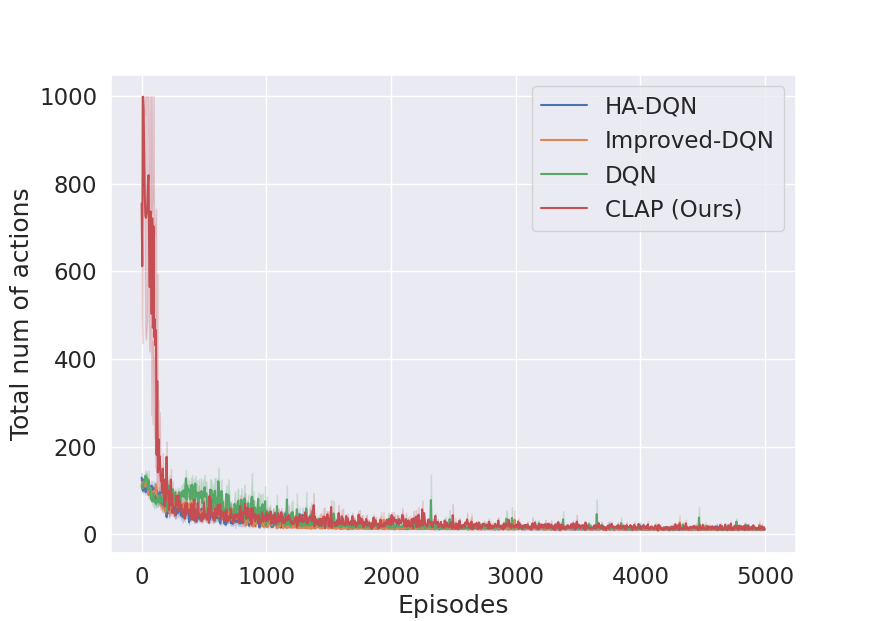}\hfil
\includegraphics[width=0.5\linewidth, height=0.32\linewidth]{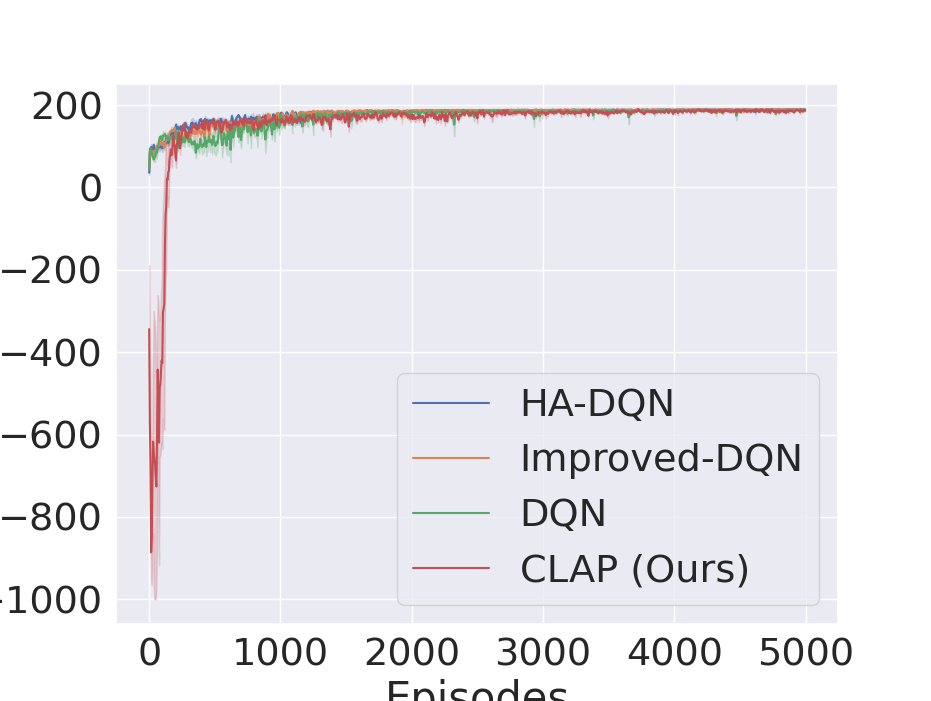}
                     }

\subfloat[Medium Network Scenario (left: Number of actions, right: Rewards)]{
\includegraphics[width=0.5\linewidth, height=0.32\linewidth]{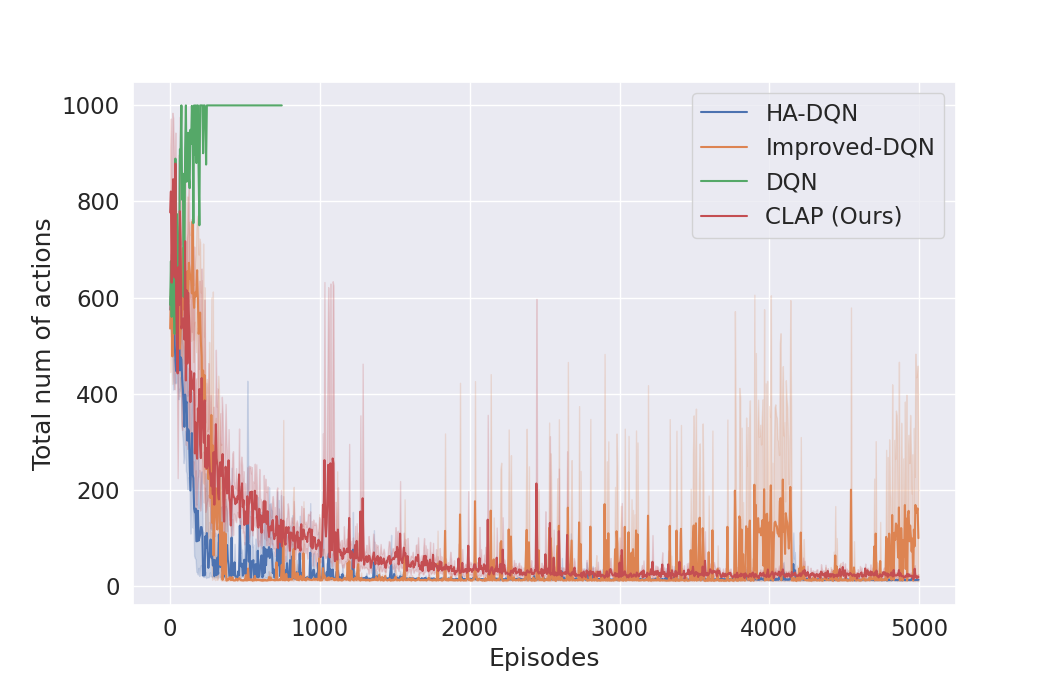}\hfil
\includegraphics[width=0.4\linewidth, height=0.32\linewidth]{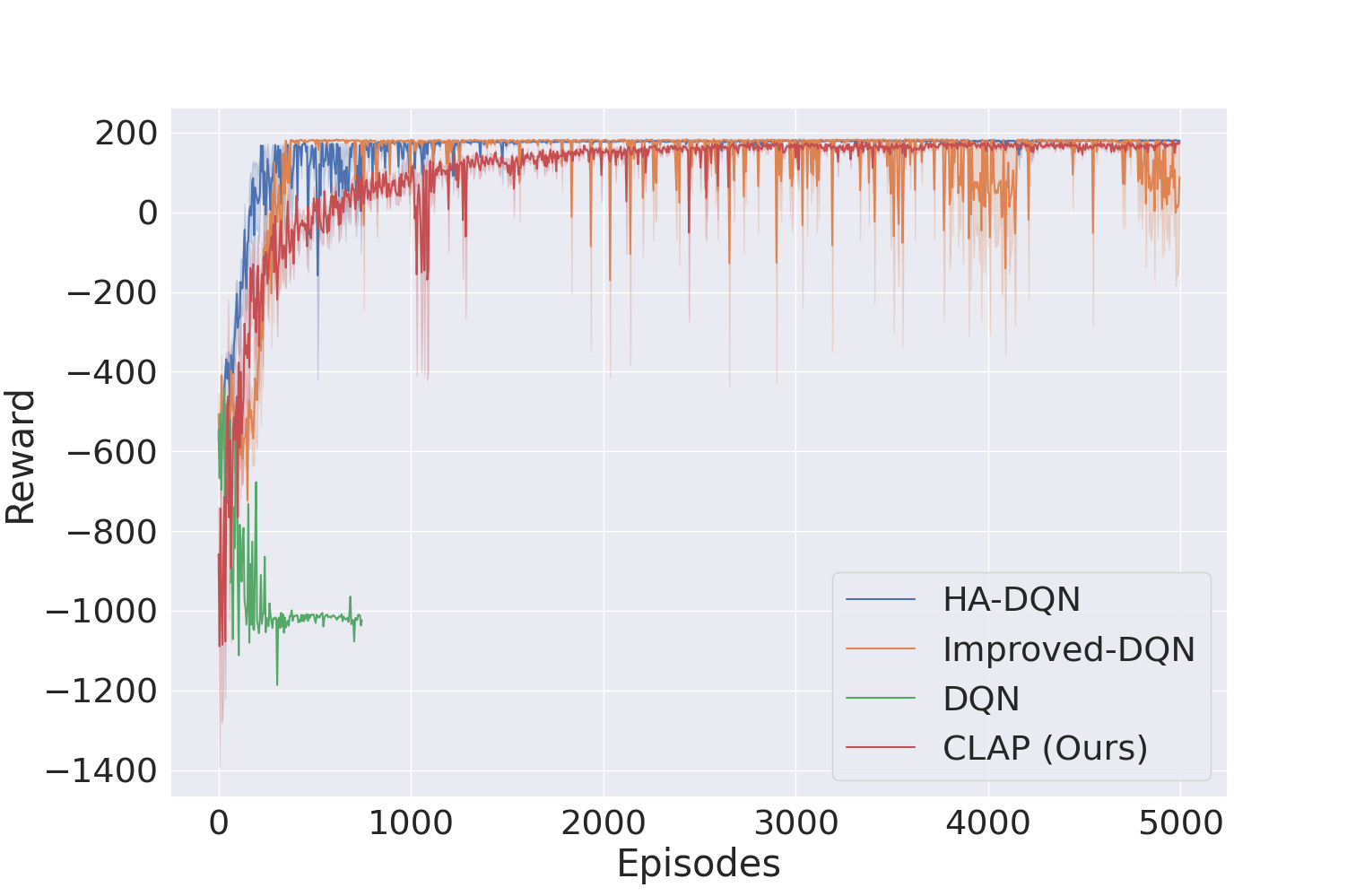}
                     }

\subfloat[Large Network Scenario (left: Number of actions, right: Rewards)]{
\includegraphics[width=0.5\linewidth, height=0.32\linewidth]{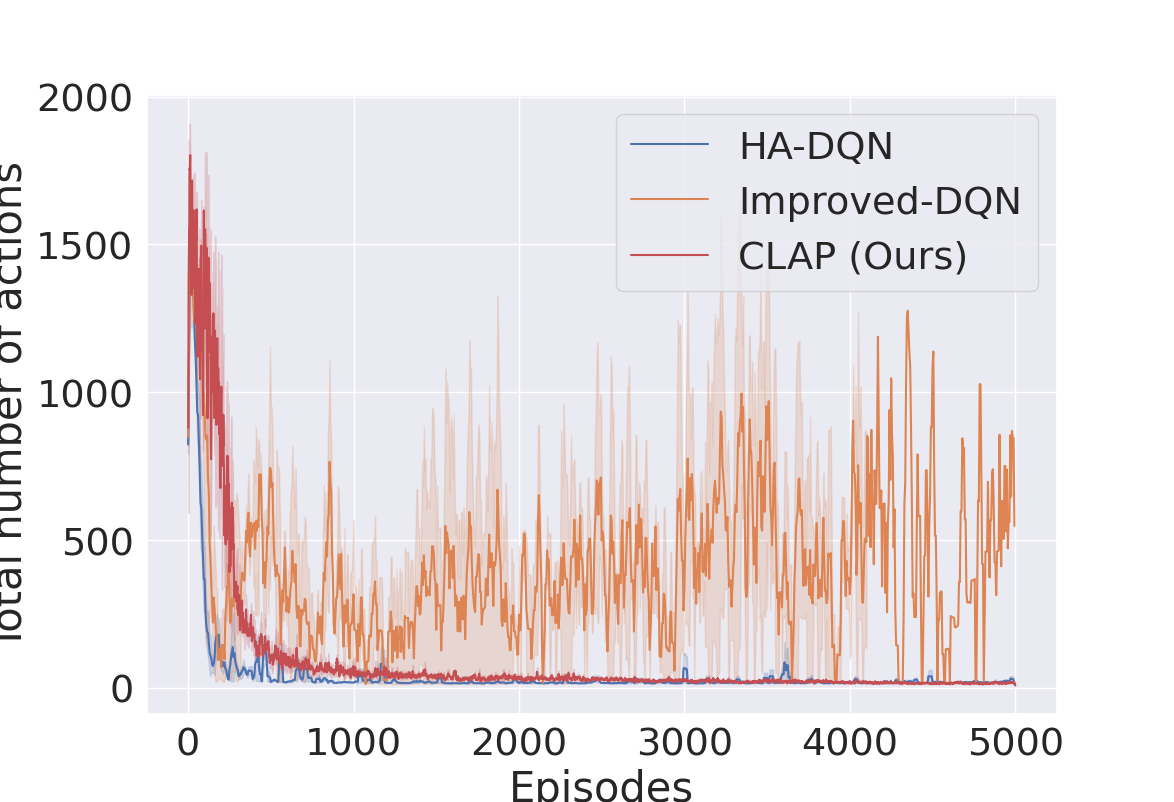}\hfil
\includegraphics[width=0.5\linewidth, height=0.32\linewidth]{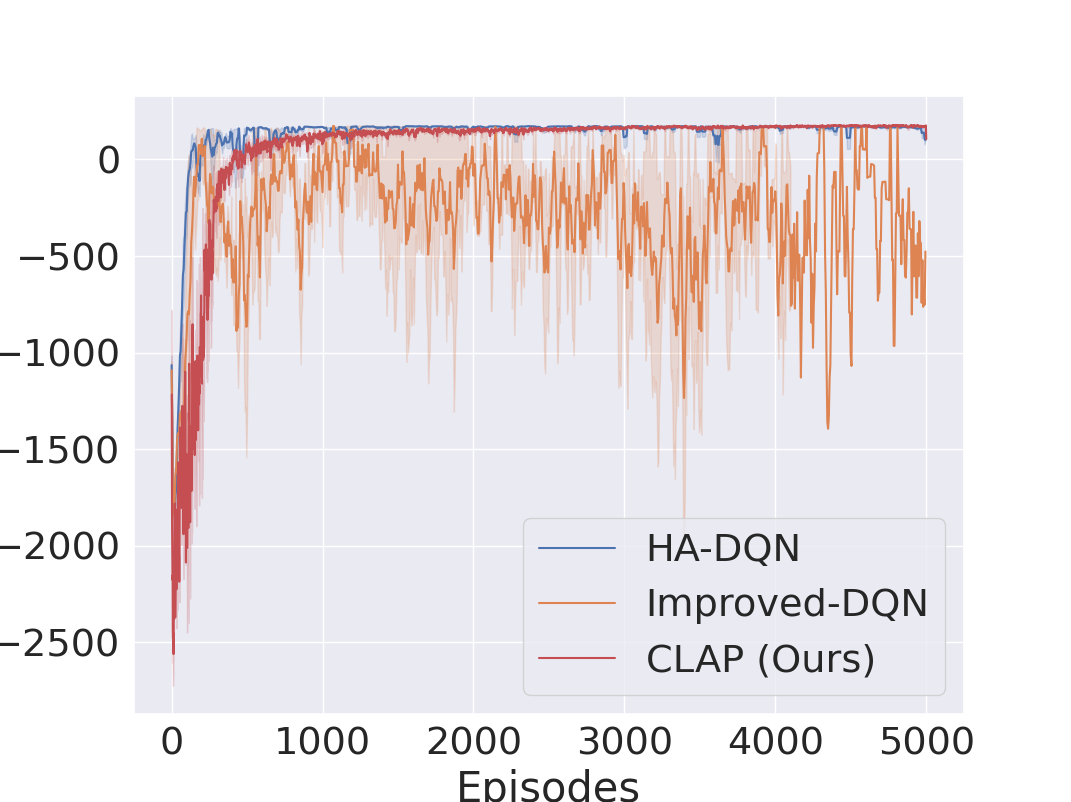}
                     }

\caption{Training performance on different NASim scenarios}\label{fig:NASIM}
\end{figure}

\paragraph{CAGE}
Compared with NASim, CAGE is designed with more sophisticated features. Firstly, CAGE implements three groups of actions: host-to-host actions, host-to-subnet actions, and on-host actions, whereas in NASim, only on-host actions are considered. The hosts in a subnet may only act upon hosts in the same subnet or adjacent subnets, which reflects the challenging pivoting process in real pen-testing. Therefore, the size of the total action space in CAGE is calculated as $|A|^{\operatorname{CAGE}} = |H| \times ((|H|-1) \times N^{\operatorname{Host-to-Host}} + |S| \times N^{\operatorname{Host-to-Subnet}} + N^{\operatorname{On-Host}})$, where $S$ is the set of subnets. A final high reward is given for compromise the target host. Notably, different from NASim, no action cost is given in CAGE, because the present of discount factor.

\subsection{Experiments}
We first compare the performance of our method with other RL pen-testing methods in NASim. Then the performance of our proposed coverage fusion mechanism will be examined by benchmarking on CAGE. Lastly, experiments are conducted under multi-objective settings, where the reward is in vector form to investigate the effectiveness of our proposed chebyshev decomposition method in generating behavior-diverse pen-testing agents.


\paragraph{Strategy Training}
To empirically validate the performance of these methods in terms of learning and convergence properties, we use NASim as our experiment environment. A variety of scenarios that cover a range of network sized and complexities with different number of hosts and subnets are  as shown in Table~\ref{tab:network}.

\begin{table}[htb!]
\begin{tabular}{|c|c|c|c|c|c|}
\hline
 & Subnets & Hosts & OS & Services & Processes \\ \hline
Small & 7 & 8 & 2 & 3 & 2 \\ \hline
Medium & 6 & 16 & 2 & 5 & 3 \\ \hline
Large & 8 & 23 & 3 & 7 & 3 \\ \hline
\end{tabular}
\caption{Network scenarios list}\label{tab:network}
\end{table}

The proposed approach is evaluated and compared with several other state-of-the-art solutions including DQN, improved DQN \cite{zhou2021autonomous}, as well as Hierarchy DQN \cite{tran2021deep} in terms of two metrics: the mean rewards over training episodes and the number of actions that the agent used in each episode.

Figure~\ref{fig:NASIM} demonstrates the learning performance of different methods under different network configurations. It can be seen that when the size of the target network is relatively small, all the methods were able to learn optimal policy. Nevertheless, for medium-sized network scenario, DQN is not able to learn and the performance of improved DQN becomes unstable. Meanwhile, value-based methods shown faster convergence while the proposed method demonstrates good stability, especially with larger network size. 

\paragraph{Coverage Mechanism}

To model the realistic pen-testing scenarios, CAGE is modified in a way that each action is assigned with a corresponding pre-condition. When the agent upon to select an action, the pre-condition of such action must be true. In the proposed method is compared with HA-DQN \footnote{Other methods are not compared due to their poor performance under these settings.} in terms of the number of used actions \footnote{The performance comparisons on reward per episode are not provided, as the cost of actions are not included in CAGE.
}.  



\begin{figure}[h]
    \centering
    \includegraphics[width=0.9\linewidth]{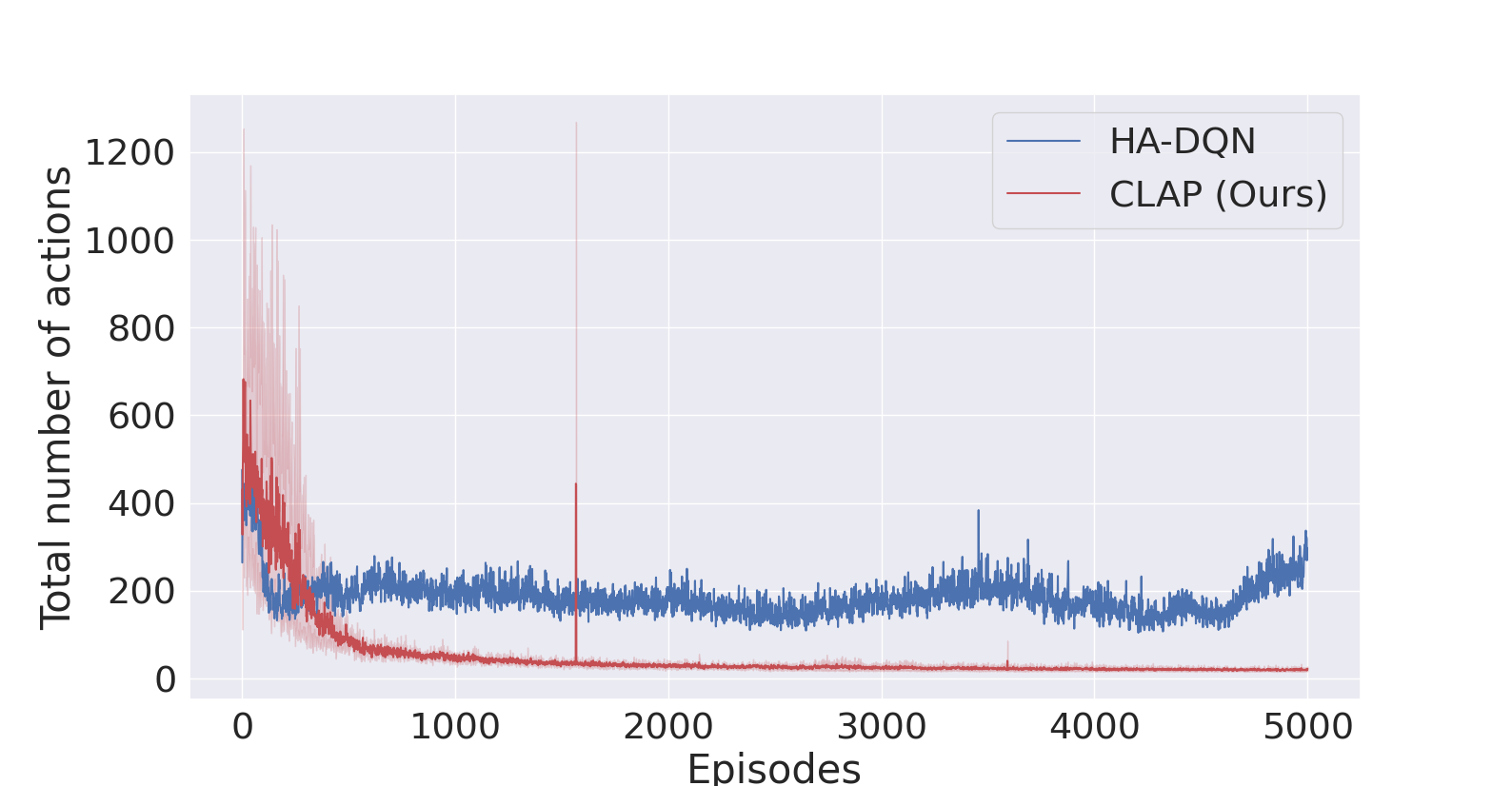}
    \caption{Number of actions per episode on CAGE}
    \label{fig:Coverage}
\end{figure}

Figure~\ref{fig:Coverage} shows the increasingly changing action space makes the training process of HA-DQN hard to converge to the optimal. Although HA-DQN can handle the large action space in CAGE, it quickly learns a sub-optimal solution and stops improving. Our proposed mechanism demonstrates strong convergence under such realistic settings, as the coverage mechanism lead the agent into continuous penetration in the target network.

\subsection{Behaviour Diversity}
We investigate the effectiveness of CLAP in generating effective attack agent in pen-testing with diverse behaviors. In this application the attack agent has to balance the effectiveness of impact the target host (also known as, capture the flag) and gather more accesses from the network. As the CAGE environment has only one reward, we further modify CAGE to have vector-formed reward function.  Apart from the reward received from compromising target hosts, our multi-objective variant adds another reward $r_{\operatorname{PrivEsc}}$ for each access the agent successfully elevated: $\mathbf{r}_t =  (r_{\operatorname{Compromise}}(s,a), r_{\operatorname{PrivEsc}}(s,a))$. The first dimension representing the value of the hosts impacted, second dimension representing the value of hosts which privilege access are gained. We also add a small negative action cost on $r_{\operatorname{Compromise}}(s,a)$. Therefore, the two optimizing  objectives of this experiment are 1) Compromise the target hosts with less actions.  2) Gather more foothold in the network which corresponding to the "depth vs breadth" dilemma in pen-testing practice. These two objectives are measured by the number of actions to compromise the target and the number of access gain in the network, respectively.

\begin{figure}
    \centering
    \includegraphics[width=0.9\linewidth]{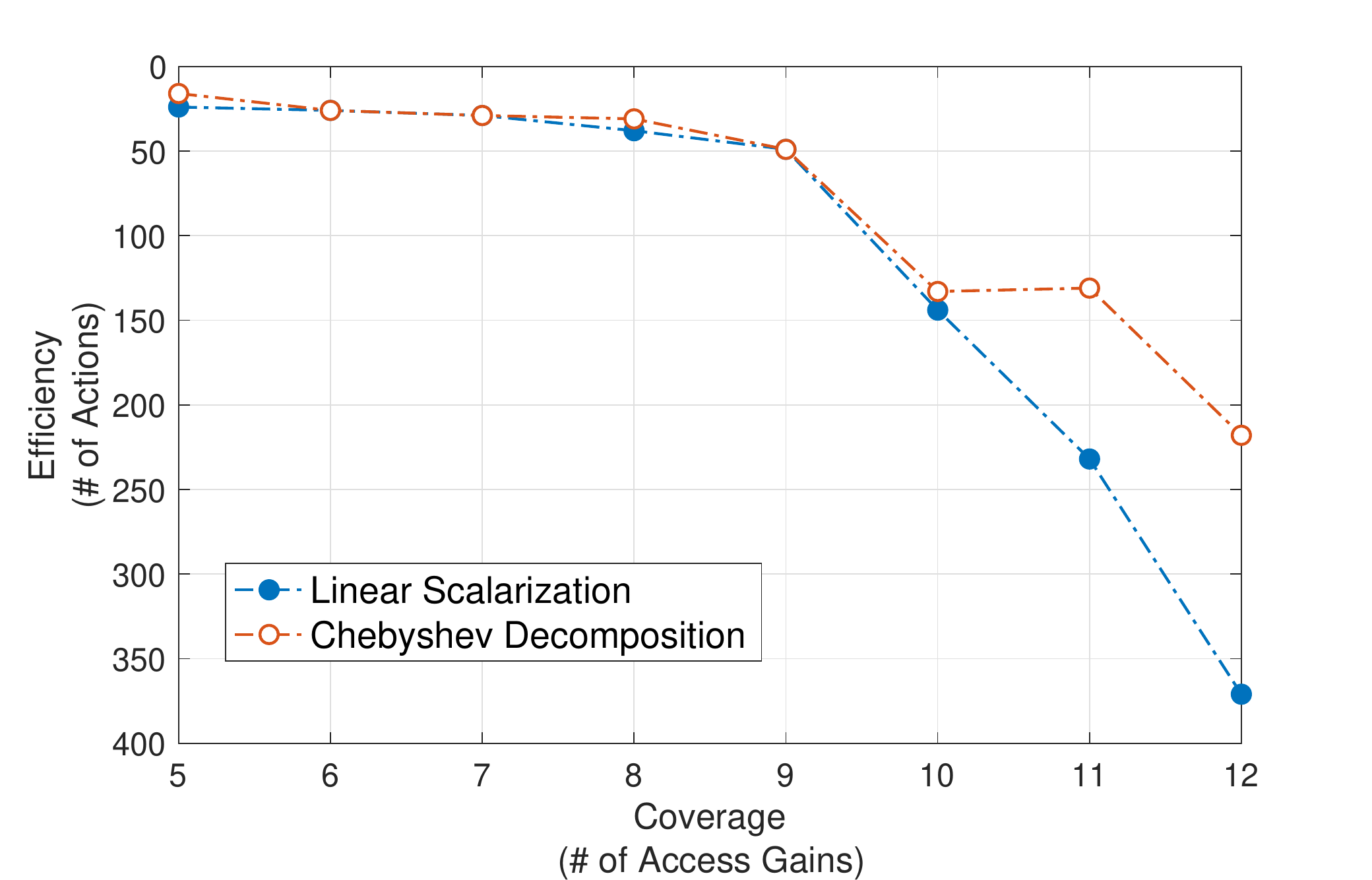}
    \caption{Distribution of attack agents generated among two objectives.}
    \label{fig:my_label}
\end{figure}

We compare out method with linear scalarization method that using reward shaping method. It transfers the multi-objective problem into single-objective learning problem by reshape the reward into a scalar by $\mathbf{r_t} = w_{\operatorname{PrivEsc}} * r_{\operatorname{PrivEsc}}(s,a) + w_{\operatorname{Compromise}} * r_{\operatorname{Compromise}}(s,a)$. Figure~\ref{fig:my_label} illustrates the distribution of attack agents created by our methods among two optimization objectives compared with the linear scalarizition method \footnote{ Different weights are uniformity selected from $[0 ,1]$ and the best performed policies are kept.}. This experiment denotes the ability of each method to discover the set of non-inferior agents, when combining the results for a range of weight settings. It can be seen that, the chebyshev critic can obtain a set of solutions that dominant a set of solution that found by linear scalatizition. The results also demonstrates our chebyshev critic's ability on handling non-concavities.


\section{Conclusion}
This paper introduces a reinforcement learning pen-testing agent that extends more generally to realistic cyber attack scenario. Additionally, CLAP generates behavior-diverse agents by leveraging chebyshev critic. Our results on pen-testing simulation platform NASim and CAGE show the effectiveness of CLAP compared with state-of-the-arts. Going forward, we believe further performance comparison with human pen-testing experts is worth investigating.

\section*{Acknowledgments}

\appendix

\bibliographystyle{named}
\bibliography{bibtex}

\end{document}